\documentclass[runningheads]{llncs}
\usepackage[T1]{fontenc}
\usepackage{graphicx}
\usepackage{booktabs}

\usepackage{tabularray}

\usepackage{algorithm}
\usepackage{algorithmic}
\graphicspath{ {./images/} }
\usepackage{physics}
\usepackage{tabularx}
\usepackage{subcaption}
\usepackage{dingbat}
\usepackage{acro}
\usepackage{amsfonts}
\usepackage{amssymb}
\usepackage{soul}

\usepackage{orcidlink}

\DeclareAcronym{MTL}{
	short=MTL,
	long=Multitask Learning,
}

\DeclareAcronym{PCGrad}{
	short=PCGrad, 
	long=Project Conflicting Gradients
}

\DeclareAcronym{DTP}{
	short=DTP, 
	long=Dynamic Task Prioritization
}

\DeclareAcronym{BEV}{
	short=BEV, 
	long=Bird's Eye View
}

\DeclareAcronym{wPCGrad}{
	short=wPCGrad, 
	long=Weighted Projection of Conflicting Gradients
}

\usepackage[capitalise]{cleveref}

\begin{document}
\title{Task Weighting through Gradient Projection for Multitask Learning} 

\titlerunning{Task Weighting through Gradient Projection for MTL}
%
\author{Christian Bohn\inst{1, 2,}\thanks{Corresponding author: \email{christian.bohn@uni-wuppertal.de}}\orcidID{0009-0006-5684-0318} \and
Ido Freeman\inst{1}\orcidID{0000-0003-4825-060X} \and
Hasan Tercan\inst{2}\orcidID{0000-0003-0080-6285} \and
Tobias Meisen\inst{2}\orcidID{0000-0002-1969-559X}}
\authorrunning{C. Bohn et al.}
%
\institute{APTIV \and
	University of Wuppertal, Wuppertal, Germany}
\maketitle              

\begin{abstract}
	In multitask learning, conflicts between task gradients are a frequent issue degrading a model's training performance. 

	This is commonly addressed by using the Gradient Projection algorithm \textit{PCGrad} that often leads to faster convergence and improved performance metrics.

	In this work, we present a method to adapt this algorithm to simultaneously also perform task prioritization. 

	Our approach differs from traditional task weighting performed by scaling task losses in that our weighting scheme applies only in cases where tasks are in conflict, but lets the training proceed unhindered otherwise. We replace task weighting factors by a probability distribution that determines which task gradients get projected in conflict cases.\\
	Our experiments on the nuScenes, CIFAR-100, and CelebA datasets 
	confirm that our approach is a practical method for task weighting.  
	Paired with multiple different task weighting schemes, we observe a significant improvement in the performance metrics of most tasks compared to Gradient Projection with uniform projection probabilities.
\end{abstract}

\section{Introduction}

\textit{\ac{MTL}} refers to training a deep neural network model to perform more than one task, thus saving compute resources and ideally leading to improved generalization, as noted by Caruana \cite{Caruana1997MultitaskL}. 

For a model to be considered a multitask network, at least one part of its architecture must be shared between all tasks, i.e., this part of the architecture is the same for all tasks. Often, these shared parts are the first layers of the network. Usually, the architecture is further divided into several task heads after these shared parts, each of them representing the predictions for a subset of the tasks. 

Training multiple tasks in a single model requires balancing tasks. Proper task weighting is necessary to avoid scenarios in which a subset of the tasks receives too much focus during training, resulting in acceptable performance on that subset, while the remaining tasks receive too little priority, which negatively impacts their performance. 

Developing an appropriate task weighting scheme is a crucial part of training any multitask machine learning model. A good choice of task weights can effectively promote generalization in the model, often making the different tasks improve each other beyond what comparable single-task models would achieve \cite{Caruana1997MultitaskL,kendall2018multitask}. 

Previously, task weighting has oftentimes been performed by applying a factor to each task loss and then summing up these weighted losses to obtain the global loss of the model \cite{ferrari_dynamic_2018,kendall2018multitask,Leang2020DynamicTW}.
There are several ways to define the weighting factors. They can be constant over the course of training, set according to a static weight schedule defined before training or set dynamically for each training iteration or epoch, based on some key performance metrics for each task \cite{ferrari_dynamic_2018}. 

Applying weight factors to the losses is a simplistic approach that does not consider possible complex relations between tasks. Such a task relationship falls into one of the following categories: A pair of tasks can either be in conflict, where an improvement in one task comes at the expense of worse performance in the other, or the tasks can be mutually supportive, where improvement in one also leads to improvement in the other. These cases can be differentiated based on the gradients $\nabla_\theta \mathcal{L}_k$ of the task losses $\mathcal{L}_k$ with respect to the model parameters $\theta$. If the gradients point into a similar direction (i.e., their dot-product is $\geq 0$ or the angle between them is no larger than 90 degrees), a step in the direction of the sum of these gradients will likely improve both tasks. On the other hand, for two conflicting gradients with a dot-product $<0$, a step in the direction of the gradient sum will most likely lead to an undesirable outcome: Either none of the tasks improve significantly or only the task with the larger gradient while performance on the other task deteriorates. This problem of task conflicts is addressed by the Gradient Projection algorithm, as will be shown in \cref{section:related_work}. 

Often, it can be helpful to perform the weighting between a pair of tasks differently depending on their current relation: In cases where the tasks are in conflict with one another, a strong weighting may be appropriate to focus on the more important task, while a more uniform weighting  can be used if the gradients are well aligned since an improvement in one task also leads to an improvement in the other. This work introduces a method for achieving that by applying the weighting to task gradients rather than losses, thus enabling the differentiation if tasks are in conflict with one another or not. \\

The main contributions of our work are the following:
\begin{itemize}
	\item We propose a novel extension of the Gradient Projection algorithm \cite{yu_gradient_2020}, by incorporating task weighting into the algorithm, called \textit{\ac{wPCGrad}}. 
	This allows for fine-grained task prioritization for cases where tasks are in conflict with one another. 
	\item We compare multiple ways of setting the task weights to be used with our extended Gradient Projection algorithm. These methods include both predefined, static weight schedules, containing the priority for each task for a given training epoch, as well as dynamic task prioritization, based on each task's metrics.
	\item We evaluate our method on the nuScenes \cite{caesar_nuscenes_2020}, CelebA \cite{liu2015faceattributes}, and CIFAR-100 \cite{Krizhevsky09} datasets. Our multitask training method achieves improved results compared to the Gradient Projection algorithm it is built upon. For the nuScenes detection score, quantifying the detection performance of the model, we observe up to a 4.6\% improvement, while the bird's-eye-view semantic segmentation task improved by up to 3.2\%. For the multi-label image classification on the CelebA dataset, the improvement is smaller, but our method outperforms the Gradient Projection algorithm there, as well. Similarly, our method also achieves improved metrics on the CIFAR-100 dataset.

\end{itemize}
Our novel training method is universally applicable to any multitask network, as long as there is a large enough number of conflicting gradients and we believe it can serve as a new baseline for such models.

\section{Related Work} \label{section:related_work}

\subsection{Gradient Projection}

The Gradient Projection algorithm for \ac{MTL} was first presented by Yu et al. in \cite{yu_gradient_2020}, referred to as the \textit{\ac{PCGrad}} algorithm. 

Gradient Projection is a common optimization method in \ac{MTL} that is used to stabilize and accelerate model training. It modifies the tasks' gradients and thus resolves conflicts where an improvement of one task leads to a deterioration of the others. This often leads to improved performance metrics of the final trained model, as was shown by Yu et al. \cite{yu_gradient_2020}.

\begin{figure}
	\centering
	\includegraphics[width=0.7\linewidth]{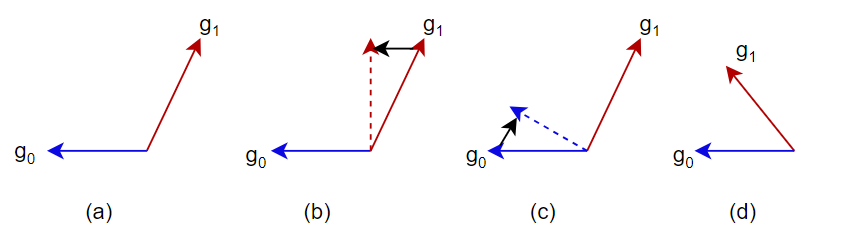}
	\caption{(a) shows a pair of gradients in conflict with each other, i.e., their dot-product is $<0$ (the angle between them is greater than 90 degrees);  To resolve this conflict, one can either project $g_1$ onto $g_0$, as in (b) or $g_0$ onto $g_1$, as in (c); (d) shows a pair of non-conflicting gradients in which case no modification is applied. (adapted from \cite{yu_gradient_2020})}
	\label{pcgrad_illustration_2_grads}
\end{figure}

It operates by resolving conflicts between tasks during training as follows: If a pair of tasks is in conflict, meaning that the dot-product between its gradients is negative, one gradient is selected uniformly at random and projected onto the normal plane defined by the other gradient in the pair, thus setting the dot-product between them to zero. This is done for each pair of gradients in the model. The concrete algorithm is presented in \cref{Algorithm_1}. Without this projection, one would usually observe slowed and degraded training since the tasks could only rarely make unhindered progress. \cref{pcgrad_illustration_2_grads} shows a schematic illustration of the algorithm's operation in all possible cases for two tasks.

\ac{PCGrad} is similar to Gradient Sign Dropout \cite{chen2020just} where one gradient in a conflicting pair is randomly dropped, i.e., set to zero, in order to resolve the conflict. Another related method is called Gradient Vaccine \cite{wang2020gradient} which, opposed to \ac{PCGrad}, is effective also when a pair of gradients share a positive cosine similarity and similarly modifies the gradients to be more closely aligned. CAGrad \cite{liu_conflict-averse_2021} introduces a second objective function to ensure that in a gradient-descent step not only the average loss across all tasks is decreased, but also every individual task loss. We chose to build our method on \ac{PCGrad} rather than other \ac{MTL} gradient balancing approaches due to that algorithm's simplicity and comparatively low computational overhead. This is in contrast to CAGrad \cite{liu_conflict-averse_2021}, for instance, where the need to optimize a second objective leads to an increase in time complexity for a single back-propagation step that is linear in the number of tasks. However, in principle, our task weighting method can also be paired with other gradient balancing methods since it only changes how gradients are sampled, not what is done to the sampled gradients (more details in \cref{method}).

\begin{algorithm} 
	\caption{\ac{PCGrad} Algorithm \cite{yu_gradient_2020}} 
	\label{Algorithm_1} 
	\begin{algorithmic}
		\REQUIRE{Model parameters $\theta$, set of tasks $\mathcal{T} = \{T_k\}$}
		\STATE $\mathbf{g}_k \leftarrow \nabla_\theta \mathcal{L}_k(\theta)$, $\forall k$ 
		\STATE $\mathbf{g}_k^{PC}\leftarrow \mathbf{g}_k$, $\forall k$
		\FOR{$T_i \in \mathcal{T}$}                    
		\FOR{$T_j \overset{\mathrm{uniformly}}{\sim} \mathcal{T}\setminus T_i$ in random order}  
		\IF{$\mathbf{g}_i^{PC} \cdot \mathbf{g}_j < 0$}  
		\STATE \textit{Subtract the projection of $\mathbf{g}_i^{PC}$ onto $\mathbf{g}_j$:}            
		\STATE Set $\mathbf{g}_i^{PC} = \mathbf{g}_i^{PC} - \frac{\mathbf{g}_i^{PC} \cdot \mathbf{g}_j}{\norm{\mathbf{g}_j}^2_2} \mathbf{g}_j$
		\ENDIF
		\ENDFOR
		\ENDFOR
		\RETURN update $\Delta\theta = \mathbf{g}^{PC} = \sum_i \mathbf{g}_i^{PC}$
		
	\end{algorithmic}
\end{algorithm}

\subsection{Dynamic Task Prioritization}

In our method, we combine the \ac{PCGrad} algorithm with task weighting following the \textit{\ac{DTP}} approach as introduced by Guo et al. in \cite{ferrari_dynamic_2018}. 
This is an optimization method for \ac{MTL} that dynamically assigns new weights to the tasks in a model for each epoch or iteration during training, based on each task’s relative performance at that time. By doing that, it effectively stabilizes the training. In \cite{ferrari_dynamic_2018}, the  dynamic weights are set based on some key performance metric for each task: This metric can be chosen from a task’s loss or some validation metric. In order to effectively balance the training, it is usually advisable to assign a higher weight to a task which is performing poorly in terms of its metric.  Each task loss is multiplied by its current weight and the losses are summed up, forming the global loss of the model. An alternative is presented in GradNorm \cite{chen_gradnorm_2018} which learns the tasks' weights through gradient descent on an auxiliary loss that encourages a scaling of the task gradients based on the rate of improvement of each of the tasks.

\subsection{BEVFormer model}

We evaluate the method presented in this work on the multitask vision transformer model \textit{BEVFormer} \cite{li_bevformer_2022} since it performs a challenging set of tasks which are related but also frequently lead to gradient conflicts: This computer vision model used in autonomous driving is trained on surround video data from six cameras to produce 3D bounding boxes and velocity vectors for a range of moving and stationary object classes present in urban traffic scenes. For each detection, the model outputs its predicted 3D bounding box, velocity vector, as well as class probabilities for 10 classes (pedestrian, bicycle, motorcycle, car, bus, construction vehicle, trailer, truck, traffic cone, barrier). The range for the detection task is $\pm51.2m$ in both x and y-direction around the ego-vehicle. Additionally it also performs \textit{Bird's Eye View Segmentation (BEV Segmentation)} of the road layout. It involves semantically segmenting the space within $\pm30m$ in x-direction and $\pm15m$ in y-direction around the ego-vehicle into three classes (road boundary, lane line, crosswalk) or background.

Its architecture comprises a convolutional backbone, extracting features from the input video frames at the current timestep. These features are then encoded into the grid-shaped \ac{BEV} queries attending both to the image features as well as recurrently to the queries of the previous timestep. The \ac{BEV} queries, forming the latent representation of the scene, are subsequently decoded by task-specific heads into the output object detections and \ac{BEV} segmentation. The backbone and \ac{BEV}-encoder are shared between the tasks, similar to \cite{xie2022m}. Notably, as opposed to previous methods, such as \cite{philion2020lift,wang2020pseudolidar}, BEVFormer does not require depth input. Previously, this model was trained by summing up the losses of the individual tasks and minimizing their sum. In this work, we replace that approach by our method.
The model is trained on the nuScenes autonomous driving dataset \cite{caesar_nuscenes_2020}, consisting of surround video data of urban traffic scenes

Recently, two successor versions of the BEVFormer model have been published: BEVFormer-v2 \cite{yang_bevformer_2022} which improves the gradient flow to the backbone and FB-BEV \cite{li2023fbbev} that reduces the number of false-positive detections by projecting both in the forward- and backward directions between the \ac{BEV} and image space. Since both of them introduce more complexity into the training of a model, we opted not to evaluate on these versions.

\section{Method} \label{method}

To combine task weighting with Gradient Projection, we introduce a probability distribution $\mathcal{D}$ that decides which task gradients get projected onto which other gradients. 
It is set up to make it more likely that the gradient of a task that is performing well already is projected onto the gradient of a relatively weaker task in case of conflict. Thus, the weaker task effectively gets prioritized.
This weighting scheme only applies if tasks are in conflict and otherwise allows each task to proceed unhindered.  

\cref{Algorithm_1} samples which gradients get projected onto which other ones uniformly at random, thus treating all tasks equally. In that case, all task prioritization can only happen during loss weighting, but not at the gradient level.  

To introduce task prioritization into gradient projection, we make the following modifications to \cref{Algorithm_1}: We adapt the algorithm to leave one task gradient completely unchanged in each optimization step and project all conflicting gradients onto it. The task whose gradient remains fully intact is sampled from the task probability distribution $\mathcal{D}$.
This means that a task with a high sampling probability is weighted more strongly. The resulting new, modified algorithm is shown in \cref{Algorithm_2}. 
The task that is sampled to remain unmodified will contribute its entire, optimal gradient to the global gradient sum, whereas all tasks conflicting with it only get to contribute a projected, suboptimal gradient.
Effectively, this leads to a prioritization of the sampled task.
The reason for the random sampling of the task gradients is that this allows for arbitrary task priorities. The alternative would be to always leave the gradient of the task with the maximum priority unchanged, which in many scenarios would lead to excessive prioritization. 

Notably, a limitation of our method is the fact that it is only suitable for multitask models with a significant number of gradient conflicts, just like PCGrad. As it is only active in case of conflicts, there will be no impact if there are no or very few conflicting gradients. 

Analogously to loss weights, the distribution $\mathcal{D}$ can be set either statically through hyperparameter tuning or dynamically with \ac{DTP}. 

\begin{algorithm} 
	\caption{\ac{wPCGrad} Algorithm} 
	\label{Algorithm_2} 
	\begin{algorithmic}
		\REQUIRE{Model parameters $\theta$, set of tasks $\mathcal{T} = \{T_k\}$}, task sampling probability distribution $\mathcal{D}$
		\STATE $\mathbf{g}_k^{PC} \leftarrow \nabla_\theta \mathcal{L}_k(\theta)$, $\forall k$ 
		\FOR{$T_i \sim \mathcal{D(}\mathcal{T})$ in random order}                    
		\FOR{$T_j \in \mathcal{T}\setminus T_i$}  
		\IF{$\mathbf{g}_i^{PC} \cdot \mathbf{g}_j^{PC} < 0$}  
		\STATE \textit{Subtract the projection of $\mathbf{g}_j^{PC}$ onto $\mathbf{g}_i^{PC}$:}            
		\STATE Set $\mathbf{g}_j^{PC} = \mathbf{g}_j^{PC} - \frac{\mathbf{g}_i^{PC} \cdot \mathbf{g}_j^{PC}}{\norm{\mathbf{g}_i}^2_2} \mathbf{g}_i$
		\ENDIF
		\ENDFOR
		\ENDFOR
		\RETURN update $\Delta\theta = \mathbf{g}^{PC} = \sum_i \mathbf{g}_i^{PC}$
		
	\end{algorithmic}
\end{algorithm}

To find a suitable task weighting scheme, we use \ac{DTP} to define a task sampling probability distribution $\mathcal{D}$ for each epoch. We set up \ac{DTP} to assign task sampling probabilities based on the share of each task loss relative to the global loss: A task with a higher loss share is assumed to perform worse and is hence assigned a higher sampling probability. The exact rule according to which the probability for task $T_i$ with loss $\mathcal{L}_i^{(t-1)}$ over the previous epoch is set for epoch $t$, is the following:
\begin{equation} \label{DTP_formula}
	p_{\mathcal{D}}(T_i) = \frac{{(\mathcal{L}_i^{(t-1)})}^\gamma}{\sum_j {(\mathcal{L}_j^{(t-1)})}^\gamma}
\end{equation}
For the focusing hyperparameter $\gamma$, similar to \cite{lin2018focal}, we found a value of $2$ to work best out of the range of values investigated for the BEVFormer model. During the first epoch, the tasks are sampled uniformly at random.  

\section{Experiments} \label{experiments_chapter}

As mentioned above, we evaluate our method on a variant of the vision transformer model BEVFormer introduced in \cite{li_bevformer_2022}. 

For our experiments, we chose a scaled-down version of the model due to compute constraints and energy usage considerations. The training time for it was roughly five days on one NVIDIA A100 GPU, while the full-scale model requires approximately an order of magnitude more compute resources. 
For details on the model architecture, please refer to \cite{li_bevformer_2022}. 

We call the first model evaluated in \cref{section:results} \textit{BEVFormer-Small}: This version of the BEVFormer architecture is similar to what is presented as BEVFormer-S in \cite{li_bevformer_2022}. The model has two output heads for both high-level tasks: object detection and \ac{BEV} segmentation. Its architecture differs in its decreased input resolution and the usage of a ResNet-50 \cite{He2015DeepRL} instead of a ResNet-101 as backbone. The backbone is pre-trained on the ImageNet \cite{5206848} dataset.  We decreased the size of the input images to 512 by 288 pixels instead of the full-scale 1600 by 900 
to further reduce the model size. We split up the loss function for this model into 4 distinct tasks: bounding-box regression, classification, and velocity estimation for each detection, as well as \ac{BEV} segmentation.  The number of parameters of this version is 43.9 million. 

Furthermore, we also evaluate our method on our own, significantly modified version of the BEVFormer architecture, called \textit{BEVFormer-Unified}. Aiming to reduce the model's size and to leverage more synergies between the tasks, this architecture performs both tasks in the same decoder. To that end, we remove the segmentation decoder from BEVFormer-Small and instead extend the set of queries passed to the remaining decoder to also cover the segmentation task, with each additional query encoding the segmentation of a patch in the BEV plane. To obtain the final BEV segmentation in the end, these patches are rearranged in a 2D grid and upsampled using a small network of deconvolutions.

For every experiment with these BEVFormer architectures, the model was trained for 24 epochs on the nuScenes training set.

The PyTorch implementation of our wPCGrad algorithm was built upon \cite{Pytorch-PCGrad} and the BEVFormer implementation we used is \cite{BEVFormer_segmentation_detection}.

We evaluate the following three ways of defining the task weights to be used with \ac{wPCGrad}: For \ac{DTP}, the task priorities are computed according to \cref{DTP_formula}, after each of the losses got multiplied with a scaling factor that is constant over time, in order for them to have roughly the same magnitude. 
For the hand-crafted prioritization schedules, we observed that bounding box regression and classification are learned well regardless of their priority, whereas velocity and BEV segmentation are the relatively more difficult tasks that become degraded if they do not receive sufficient weight.
The first such prioritization schedule (Seg. $\rightarrow$ Vel.) is set up to initially resolve almost every gradient conflict in favor of the segmentation task, until epoch 15. During the next three epochs, the tasks to be projected are sampled uniformly, and for the remaining epochs until epoch 24, the velocity task is favored. We chose the duration of the first training stage to be the longest since the model still has to learn underlying low-level features at that time. The other hand-crafted prioritization scheme (Vel. $\rightarrow$ Seg.) swaps when the segmentation and velocity tasks are being favored.

In our evaluation, we use the following metrics, based on the nuScenes validation set: Object detection and its sub-tasks is quantified by the mean Average Precision (mAP) and the true-positive metrics mean Average Attribute Error (mAAE), mean Average Orientation Error (mAOE), mean Average Scale Error (mASE), mean Average Translation Error (mATE), and mean Average Velocity Error (mAVE). To be able to measure detection quality in a single score, we report the composite nuScenes Detection Score (NDS) defined as follows: 
\begin{equation}\label{nds_equation}
	NDS = \frac{1}{10}[5\cdot mAP + \sum_{mTP \in \mathbb{TP}} (1 - \min(1, mTP))]
\end{equation}
where $\mathbb{TP}$ refers to the set of the five true-positive metrics defined above. 
For the segmentation task, we measured the mean Intersection-over-Union (mIoU) between the prediction and ground-truth segmentation map. For more details about these metrics, refer to Sec. 3.1 in \cite{caesar_nuscenes_2020}. 

We also conducted experiments on the CIFAR-100 dataset: For this evaluation we trained a Routing Network, as presented in \cite{rosenbaum2017routing}, with one output head per task. We report the average accuracy across the 20 tasks. 

For our evaluation on the CelebA dataset \cite{liu2015faceattributes}, we trained a variant of the model architecture presented in \cite{NEURIPS2018_432aca3a} on it and report the average accuracy over the 40 tasks in the dataset. 

\section{Results} \label{section:results}
\subsection{Multitask Training without \ac{wPCGrad}} \cref{table:single_task_comparison} shows that for BEVFormer-Small, the \ac{MTL} approach without \ac{PCGrad} does not lead to an improvement of all tasks compared to its single-task versions. This confirms the claim made in \cite{ma2023visioncentric} that the joint training of \ac{BEV} segmentation and object detection usually does not lead to an improvement compared to single-task models. In fact, the segmentation task works best in the single task model. However, the usage of \ac{PCGrad} during training significantly improves the performance of the multi-task model, especially for the detection task, where it leads to an improvement over the respective single-task model.

\begin{table}[H]
	\centering
	\caption{Comparison of BEVFormer-Small to its single-task versions in terms of nuScenes Detection Score (NDS) and \ac{BEV} segmentation mIoU on the nuScenes validation set. The models shown in the first two rows are trained only on the segmentation and detection task, respectively, with the loss for the other task set to zero. The row labeled Multitask shows the metrics for the BEVFormer trained on both tasks, and the final row shows the impact of training the multitask model using PCGrad. }
	\resizebox{0.4\textwidth}{!}{
		\begin{tblr}{l l l} 
			\hline[1pt]
			&NDS $\uparrow$	&mIoU $\uparrow$  	\\[0.5ex] 
			\hline 
			Seg. only			&- 				&\textbf{0.385} 	\\ 
			Det. only  			&0.323	&- 			 		\\
			Multitask			&0.317 			&0.336 				\\ 
			Multitask+PCGrad  &\textbf{0.329}	&0.345 				\\
			\hline[1pt]
		\end{tblr}
	}
	\\ [1ex]	
	\label{table:single_task_comparison}
\end{table}

\subsection{Impact of \ac{wPCGrad}}

\begin{table}[htb]
	\centering
	\caption{Comparison of the training methods investigated across the relevant metrics for all tasks on the nuScenes validation set. We compare BEVFormer-Small to the same model trained with PCGrad and three weighting schemes for wPCGrad: The handcrafted static prioritization schedule favoring the segmentation task first followed by the velocity task in the later epochs, the static schedule with velocity first followed by segmentation, and the DTP setup with $\gamma = 2$. Improvement percentages are reported relative to the metrics achieved with PCGrad. All scores  are averaged over two training runs with different random seeds.
	}
	\resizebox{\textwidth}{!}{
		\begin{tblr}{l l l|l l l l l|l}
			\hline[1pt]
			&NDS $\uparrow$&mAP $\uparrow$&mAAE $\downarrow$&mAOE $\downarrow$&mASE $\downarrow$&mATE $\downarrow$&mAVE $\downarrow$&mIoU $\uparrow$ \\[0.5ex]
			\hline
			
			BEVFormer-Small \cite{li_bevformer_2022}	&0.317			&0.186 			&0.215 			&0.704			&0.300 			&0.895 &0.648 			&0.336 	\\ 
			
			$\rotatebox[origin=c]{180}{$\Lsh$}$ + PCGrad \cite{yu_gradient_2020}	&0.329		&0.195			&0.212	&0.698			&\textbf{0.297}	&0.898			&0.579			&0.345	\\
			
			$\rotatebox[origin=c]{180}{$\Lsh$}$ + wPCGrad  Seg. $\rightarrow$ Vel. (ours)&0.341 +3.6\%			&0.207	+6.2\%	&0.213			&0.667			&\textbf{0.297}	&0.904			&0.549	&0.355	+2.9\%	\\
			
			$\rotatebox[origin=c]{180}{$\Lsh$}$ + wPCGrad  Vel. $\rightarrow$ Seg. (ours)&0.339 +3.0\%	&0.203	+4.1\%	&\textbf{0.203}	&0.690	&0.299	&\textbf{0.891}			&\textbf{0.544}	&\textbf{0.356} +3.2\%	\\
			
			$\rotatebox[origin=c]{180}{$\Lsh$}$ + wPCGrad DTP (ours) &\textbf{0.344} +4.6\% &\textbf{0.209} +7.2\% &0.214 &\textbf{0.635} &0.302 &0.901 &0.559 &0.352 +2.0\% \\
			\hline[1pt]
		\end{tblr}
	}
	\\ [1ex]
	\label{table:ablation_study}
\end{table}

\cref{table:ablation_study} shows that applying the \ac{PCGrad} optimization to the baseline BEV-\ Former-Small already leads to a significant improvement in most metrics. This confirms the results shown in \cite{yu_gradient_2020}, where applying this method also yielded similar improvements across a wide variety of models and datasets. The addition of task weighting with \ac{wPCGrad} using either the handcrafted schedules or \ac{DTP} leads to a comparable additional improvement over \ac{PCGrad} across most metrics, demonstrating the efficacy of our method. In fact, for all metrics investigated, the best scores are achieved with one of the \ac{wPCGrad} setups. 

We observe the most significant improvement in the mAP score for the detection task which in turn is also reflected in the NDS score. Across the true-positive metrics, the improvements are not universally as large, however the usage of \ac{wPCGrad} still has a positive impact. 

For the segmentation mIoU metric, the improvement is slightly smaller than for the detection task, however still relevant, as will also be evident in our qualitative results in \cref{subsection:qualitative_results}.

\begin{figure}
	\centering
	\begin{subfigure}{0.24\linewidth}
		\includegraphics[width=\textwidth]{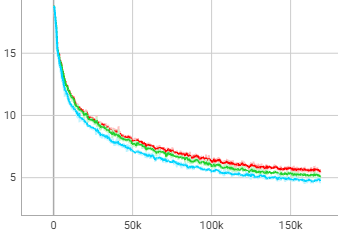} 
		\caption{Training Loss}
		\label{metrics_curves_imgs:loss}
	\end{subfigure}
	\begin{subfigure}{0.24\linewidth}
		\includegraphics[width=\textwidth]{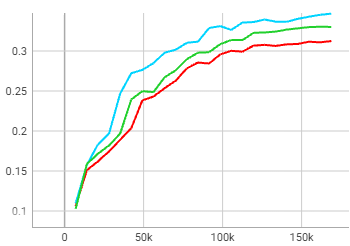} 
		\caption{NDS}
		\label{metrics_curves_imgs:nds}
	\end{subfigure}
	\begin{subfigure}{0.24\linewidth}
		\includegraphics[width=\textwidth]{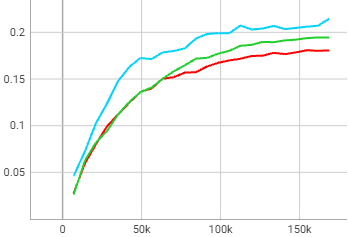} 
		\caption{mAP}
		\label{metrics_curves_imgs:map}
	\end{subfigure}
	\begin{subfigure}{0.24\linewidth}
		\includegraphics[width=\textwidth]{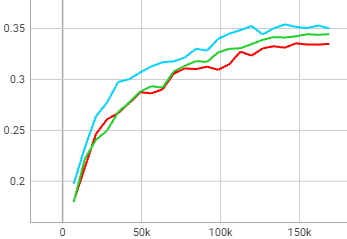} 
		\caption{mIoU}
		\label{metrics_curves_imgs:miou}
	\end{subfigure}
	\caption{Training Loss of BEVFormer-Small on the nuScenes training set, NDS, mAP, and mIoU  on the validation set for the baseline setup without PCGrad (red), with PCGrad (green), ours with Dynamic Task Prioritization (blue).}
	\label{metrics_curves_imgs}
\end{figure}

\cref{metrics_curves_imgs} shows the most important metrics over the course of training for BEVFormer-Small without Gradient Projection, with \ac{PCGrad}, and the model trained using \ac{wPCGrad} with \ac{DTP}. For each additional feature, we observe an improvement compared to the preceding setup. 

In \cref{table:ablation_study_bevformer_unified}, we show the metrics for BEVFormer-Unified, once trained with PCGrad and once with wPCGrad using a dynamic task sampling distribution from DTP. Similarly to the results in \cref{table:ablation_study}, comparing our method to PCGrad, we observe significant improvements in most metrics, albeit with slightly smaller magnitudes. Note that the segmentation mIoU score is not comparable to the ones in \cref{table:single_task_comparison} and \cref{table:ablation_study}  and expected to be lower since the perception range for the segmentation task is much larger in this experiment, in order to match the range for the detection task. 

\begin{table}[htb]
	\centering
	\caption{Results on BEVFormer-Unified. We compare the model trained with PCGrad to the one trained using wPCGrad with DTP for $\gamma = 2$. 
	}
	\resizebox{\textwidth}{!}{
		\begin{tblr}{l l l|l l l l l|l}
			\hline[1pt]
			&NDS $\uparrow$&mAP $\uparrow$&mAAE $\downarrow$&mAOE $\downarrow$&mASE $\downarrow$&mATE $\downarrow$&mAVE $\downarrow$&mIoU $\uparrow$ \\[0.5ex]
			\hline
			
			BEVFormer-Unified + PCGrad \cite{yu_gradient_2020}		&0.305		&0.181		&0.224			&\textbf{0.739}	&0.315			&0.930			&0.646	&\textbf{0.185}\\ 
			
			BEVFormer-Unified + wPCGrad DTP (ours) &\textbf{0.312}		&\textbf{0.189}	&\textbf{0.218}			&0.756	&\textbf{0.313}			&\textbf{0.910}			&\textbf{0.630}	&0.183  \\
			 
			\hline[1pt]
		\end{tblr}
	}
	\\ [1ex]
	\label{table:ablation_study_bevformer_unified}
\end{table}

\subsection{Evaluation on CelebA dataset}

As mentioned before, we also evaluate our method on the multi-label classification problem posed by the CelebA dataset \cite{liu2015faceattributes}. It contains 200K images of faces, with 40 attribute annotations each. We consider the prediction of each of these attributes as a task. As a baseline model architecture, we use the implementation provided by Sener and Koltun \cite{NEURIPS2018_432aca3a}. Since we observed strong overfitting with the base implementation, we decreased the number of channels used in the feature extraction backbone. The task weighting method used with wPCGrad in this case was DTP as in \cref{DTP_formula} with $\gamma = 4$. 

Similarly to our results using the BEVFormer architecture on the nuScenes dataset, in \cref{table:celeba_results}, we observe an improvement in the model's accuracy with \ac{wPCGrad} compared to PCGrad. Due to a smaller relative number of gradient conflicts in this case, the improvement is smaller than for our BEVFormer experiments. 

\begin{table}
	\centering
	\caption{Average Accuracy over the 40 tasks in the CelebA dataset}
	\resizebox{0.5\textwidth}{!}{
		\begin{tblr}{l l} 
			\hline[1pt]
			& Avg. Accuracy $\uparrow$	\\[0.5ex] 
			\hline
			Baseline model	\cite{NEURIPS2018_432aca3a}				&0.844			\\
			$\rotatebox[origin=c]{180}{$\Lsh$}$ + PCGrad \cite{yu_gradient_2020}					&0.846			\\
			$\rotatebox[origin=c]{180}{$\Lsh$}$ + wPCGrad DTP (ours)		&\textbf{0.850}			\\ 
			\hline[1pt]
		\end{tblr}
	}
	\\ [1ex]
	\label{table:celeba_results}
\end{table}

\subsection{Evaluation on CIFAR-100 dataset}
As outlined above, we conducted experiments on the CIFAR-100 dataset, as well. In that dataset, each image belongs to one of 100 object classes which themselves belong to 20 super-classes. Our experiment considers the image classification for each of these super-classes as a separate task, resulting in a 20-task model. For this evaluation we trained a Routing Network, as presented by Rosenbaum et al. in \cite{rosenbaum2017routing}, with one output head per task. For the focusing hyperparameter $\gamma$ needed in DTP, we used a value of $1$ in this case. \\
As in the evaluations above, \cref{table:cifar_results} shows an improvement in the accuracy of the model when replacing PCGrad with wPCGrad paired with Dynamic Task Prioritization. 

\begin{table}
	\centering
	\caption{Average Accuracy over the 20 tasks in the CIFAR-100 test set}
	\resizebox{0.65\textwidth}{!}{
		\begin{tblr}{l l} 
			\hline[1pt]
			& Avg. Accuracy $\uparrow$	\\[0.5ex] 
			\hline
			Routing Network + PCGrad 	\cite{yu_gradient_2020}		&0.615			\\
			Routing Network + wPCGrad DTP (ours)					&\textbf{0.624} \\
			\hline[1pt]
		\end{tblr}
	}
	\\ [1ex]
	\label{table:cifar_results}
\end{table}

\subsection{Qualitative results on nuScenes dataset} \label{subsection:qualitative_results}
\begin{figure}[H]
	\centering
	\begin{subfigure}{\linewidth}
		\parbox[b][5cm][c]{0.05\textwidth}{
			\caption{}
			\label{img:qualitative-results-imgs-a}
		}
		\parbox[b]{0.95\textwidth}{
			\rotatebox[origin=l]{90}{
				\ \ \ PCGrad
			} \includegraphics[width=0.975\linewidth]{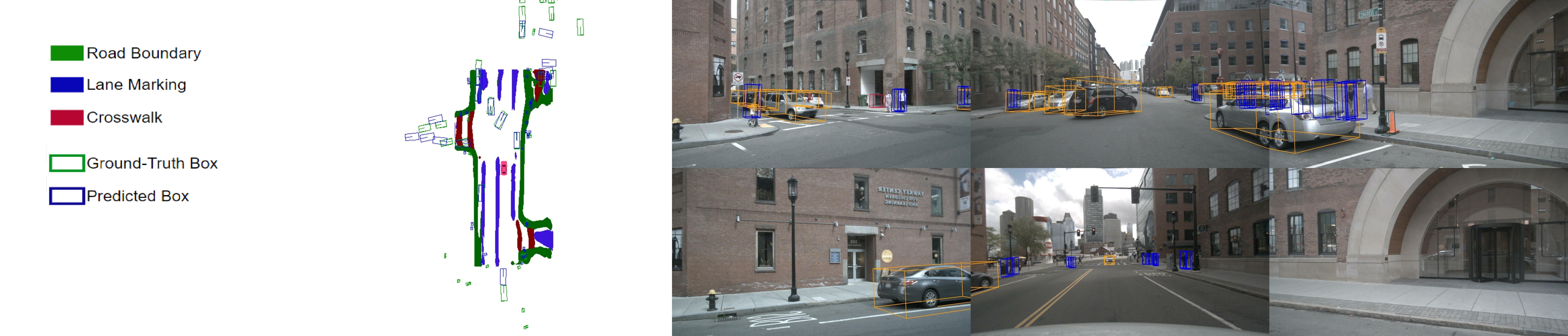}
			\rotatebox[origin=l]{90}{
				\ \ \ \ \ \ Ours
			}
			\includegraphics[width=0.975\linewidth]{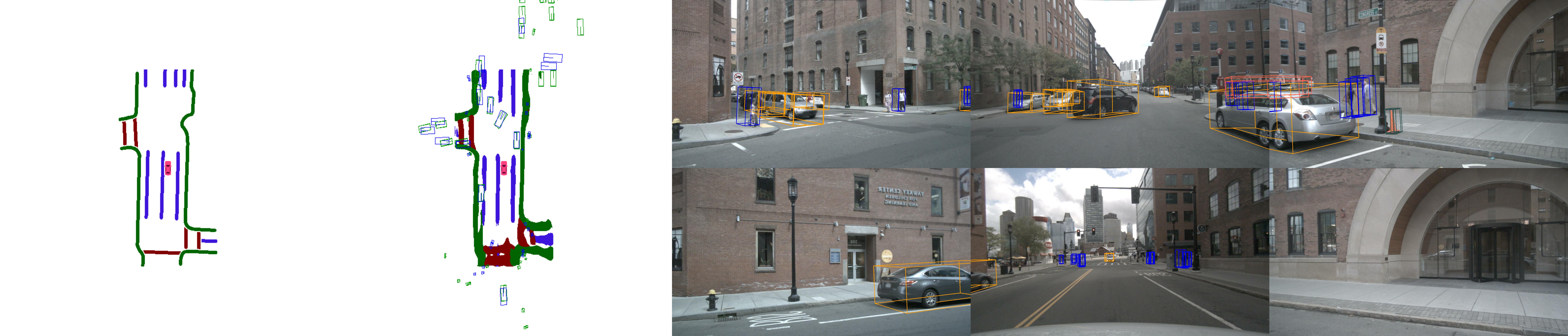}
		}
		\vspace*{0.1cm}
	\end{subfigure}
	\begin{subfigure}{\linewidth}
		\parbox[b][5cm][c]{0.05\textwidth}{
			\caption{}
			\label{img:qualitative-results-imgs-b}
		}
		\parbox[b]{0.95\textwidth}{
			\rotatebox[origin=l]{90}{
				\ \ \ PCGrad
			}
			\includegraphics[width=0.975\linewidth]{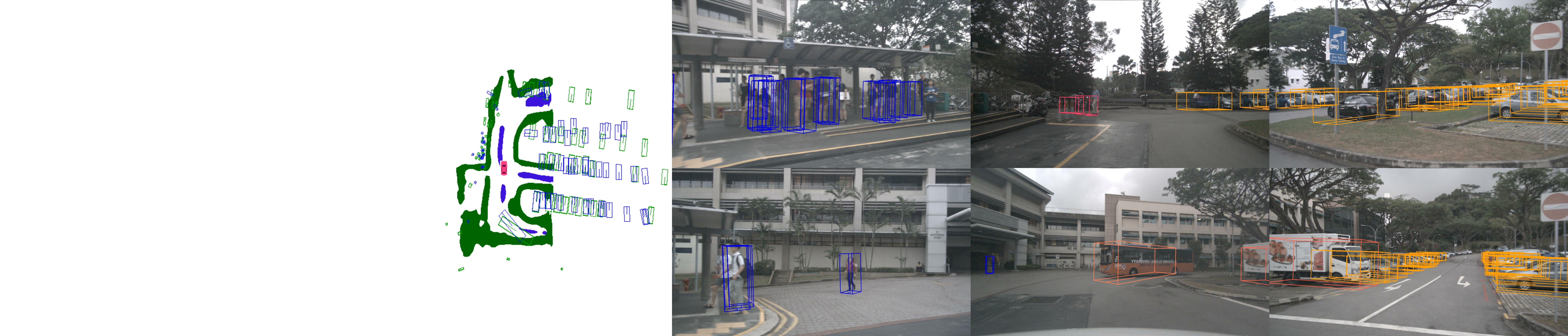}
			\rotatebox[origin=l]{90}{
				\ \ \ \ \ \ Ours
			}
			\includegraphics[width=0.975\linewidth]{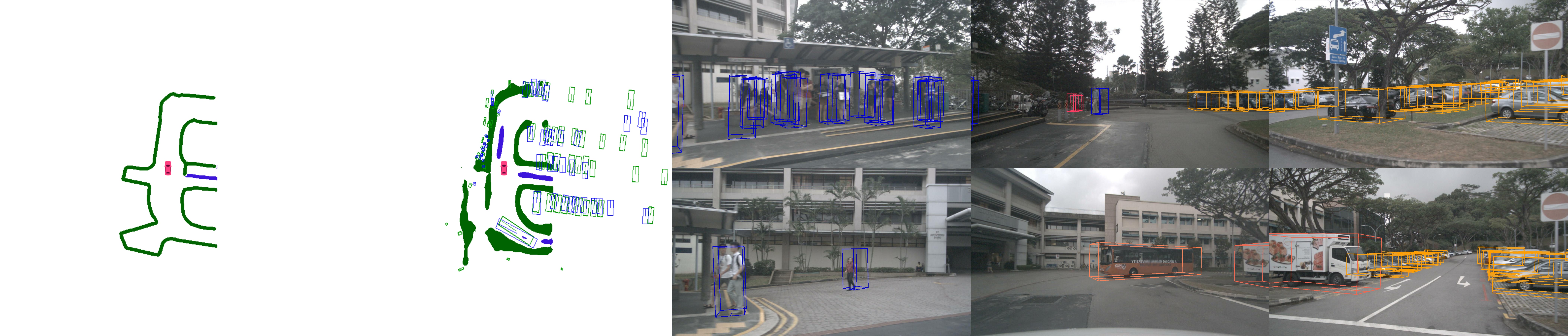}
		}
	\end{subfigure}
	\caption{Qualitative examples on the nuScenes validation set: The leftmost images show the ground-truth \ac{BEV} segmentation with the ego-vehicle in the center, next to that the predicted \ac{BEV} segmentation can be seen with ground-truth and predicted bounding boxes overlaid, and to the right the predicted bounding boxes are shown reprojected into the input images.} 
	\label{img:qualitative-results-imgs}
\end{figure}

\cref{img:qualitative-results-imgs-a} and \cref{img:qualitative-results-imgs-b} show two representative qualitative examples of the improvement from adapting projection probabilities on the nuScenes validation set. For each example, the images on top are the output from BEVFormer-Small trained with \ac{PCGrad} as in \cref{Algorithm_1} and the images below are produced by the same model trained with \ac{wPCGrad} and projection probabilities favoring the \ac{BEV} segmentation task in the early epochs followed by the velocity task in the later ones.

In general, it can be seen that the \ac{BEV} segmentation maps produced with \cref{Algorithm_2} better match the ground-truth. For the object detection task, there are significantly fewer false positives visible, reflecting the improvement in mAP shown in \cref{table:ablation_study} and the quality of the true positive detections is improved, as well. 

In particular, for \cref{img:qualitative-results-imgs-a}, note the improved segmentation of the road edge in the top-right corner of the \ac{BEV} map, as well as the correct segmentation of the crosswalk at the bottom. Also the improvement in bounding box accuracy for the vehicles in the side street to the left is apparent.  
For the parking-lot example \cref{img:qualitative-results-imgs-b} below, the reconstruction from the model trained with \cref{Algorithm_2} is more coherent than the one above. Particularly, note the improvement in road-edge segmentation for the curbs on the right side and the general improved bounding box quality for the visible parked cars. For the the parked cars that are not visible to the camera, we can observe a reduction in false positive detections. 

\section{Conclusion}
In this work, we introduced \ac{wPCGrad}, an extension to the established \ac{PCGrad} algorithm that additionally takes the priority of a task into account and adjusts which task gradients to project accordingly. Our method allows a differentiation in task weighting depending on if tasks are in conflict with one another or not: In case of conflicts, we apply a strong task weighting, whereas otherwise we effectively use uniform priorities letting each task's training proceed unhindered. This was not previously possible using traditional loss scaling to prioritize tasks. 
We investigated different task prioritization strategies and evaluated them on our algorithm.
We showed experimentally for four models trained on the nuScenes, CelebA, and CIFAR-100 datasets that our method is a significant improvement over the previous PCGrad algorithm.

\begin{credits}

\subsubsection{\discintname}
The authors have no competing interests to declare that are
relevant to the content of this article.
\end{credits}
%
%
\bibliographystyle{splncs04}
\bibliography{multi-task_learning_research}

\end{document}